\documentclass[wcp]{jmlr}

\usepackage{amsmath,amssymb,graphicx,url}

\usepackage{longtable}

\usepackage{booktabs}

\usepackage{multirow}

\jmlrvolume{266}
\jmlryear{2025}
\jmlrworkshop{Conformal and Probabilistic Prediction with Applications}
\jmlrproceedings{PMLR}{Proceedings of Machine Learning Research}

\title{Conformal Risk Control for Pulmonary Nodule Detection}
\author{\Name{Roel Hulsman}\footnote{Corresponding author. Work performed while affiliated with the JRC, before moving to Amsterdam. \\
Code is available at \url{https://github.com/roelhulsman/ConformalNoduleDetection}.}\Email{r.p.hulsman@uva.nl}\\
\addr{University of Amsterdam, Amsterdam, Netherlands}
\AND
\Name{Valentin Comte}\Email{valentin.comte@ec.europa.eu}
\\
\Name{Lorenzo Bertolini}\Email{lorenzo.bertolini@ec.europa.eu}
\\
\Name{Tobias Wiesenthal}\Email{tobias.wiesenthal@ec.europa.eu}
\\
\Name{Antonio Puertas Gallardo}\Email{antonio.puertas-gallardo@ec.europa.eu}
\\
\Name{Mario Ceresa}\Email{mario.ceresa@ec.europa.eu}
\\
\addr{European Commission, Joint Research Centre (JRC), Ispra, Italy}}
\editor{Khuong An Nguyen, Zhiyuan Luo, Tuwe Löfström, Lars Carlsson and Henrik Boström}

\begin{document}

\maketitle

\begin{abstract}
Quantitative tools are increasingly appealing for decision support in healthcare, driven by the growing capabilities of advanced AI systems. However, understanding the predictive uncertainties surrounding a tool's output is crucial for decision-makers to ensure reliable and transparent decisions. In this paper, we present a case study on pulmonary nodule detection for lung cancer screening, enhancing an advanced detection model with an uncertainty quantification technique called conformal risk control (CRC). We demonstrate that prediction sets with conformal guarantees are attractive measures of predictive uncertainty in the safety-critical healthcare domain, allowing end-users to achieve arbitrary validity by trading off false positives and providing formal statistical guarantees on model performance. Among ground-truth nodules annotated by at least three radiologists, our model achieves a sensitivity that is competitive with that generally achieved by individual radiologists, with a slight increase in false positives. Furthermore, we illustrate the risks of using off-the-shelve prediction models when faced with ontological uncertainty, such as when radiologists disagree on what constitutes the ground truth on pulmonary nodules.
\end{abstract}


\section{Introduction}
Lung cancer is the leading cause of cancer-related deaths worldwide \citep{torre2016lung}. To improve survival rates, low-dose thoracic CT screening offers potential for the early detection of high-risk individuals \citep{national2011reduced}. In this context, AI-assisted detection of malignant nodules could serve as a meaningful instrument to implement CT screening on a larger scale. In fact, a recent survey among radiologists in the United States identified lesion detection as the top answer to what they would like AI to do to enhance their clinical practices \citep{allen20212020}, while also highlighting the need for methods to evaluate and report the performance of AI systems on representative datasets. \citet{hendrix2023deep} presents a state-of-the-art example of such a system, showing reliable detection of benign and malignant pulmonary nodules in clinically indicated CT scans. 

A trustworthy representation of predictive uncertainty is crucial to evaluating the performance of AI systems, and it should be regarded a key feature in safety-critical healthcare applications \citep{begoli2019need, chua2023tackling}. In the case of AI-assisted pulmonary nodule detection, a performance assessment measures sensitivity, outlining the number of undetected nodules (that is, false negatives), as well as precision, constraining the number of false positives. A system that too often fails to detect nodules cannot be considered a safe system for large-scale clinical use, while a system that returns too many false positives has limited practical benefit for radiologists. 

Common practice in AI-assisted pulmonary nodule detection is to evaluate the performance of an AI system using free-response receiver operating characteristic (FROC) analysis \citep{chakraborty2013brief, van2010comparing, gu2021survey}, where the sensitivity of an AI system on a particular dataset is offset against the average number of false positives per scan. Each point on a FROC plot is determined by a confidence threshold $\lambda$, indicating the degree of confidence for a system to distinguish nodules from non-nodules. Typically, an AI systems in clinical use has their internal threshold set to operate somewhere between 1 and 4 false positives per scan on average \citep{van2010comparing}. Since pulmonary nodule detection models tend to rely on neural network architectures, they generally do not provide performance guarantees on the sensitivity per scan. 

To increase trust in AI-based decision support in high-risk domains, valid statistical inference is particularly important. To ensure a patient-centric detection model operates with a high sensitivity per scan, we explore the added benefit of a predictive uncertainty quantification technique from the conformal prediction framework \citep{vovk2005algorithmic, angelopoulos2023conformal}, specifically a novel method called conformal risk control (CRC) \citep{angelopoulos2024crc}. The conformal prediction framework relates predictive uncertainty to the size of prediction sets, imitating the human tendency of signalling uncertainty by offering alternatives when unsure \citep{cresswell2024conformal}. Conformal methods\footnote{Technically, we refer specifically to the popular version of \emph{split} conformal prediction \citep{papadopoulos2002inductive, lei2018distribution} and its extensions to distribution-free risk control \citep{bates2021distribution, angelopoulos2021learn, angelopoulos2024crc}, but for simplicity we use the umbrella term conformal prediction.} wrap around an arbitrary prediction model to turn point predictions into prediction sets with formal statistical guarantees on a particular risk function, such as the sensitivity per scan. 

Assuming access to a calibration dataset representative\footnote{The formal statistical requirement is for future scans to be \emph{exchangeable} with the calibration data.} of future scans, we obtain interpretable statements for non-technical radiologists, such as: \emph{``The calibrated detection model is expected to miss at most $x\%$ of ground-truth nodules per scan, on average over future scans.''}. This guarantee holds in expectation, and it could be informative to decision-makers that operate on the population level. For example, to manage downstream diagnostics or to allocate treatment resources. For the purposes of this case study, CRC amounts to a mathematically principled alternative to FROC analysis. 

Finally, the task of lung nodule detection in CT scans is consensus-driven, requiring radiologists to agree on the distinction between nodules and non-nodules. Substantial variability exists across radiologists in such a task, even among experienced thoracic radiologists \citep{armato2009assessment}. Disagreement among experts on what constitutes the ground truth concerning a pulmonary nodule is a matter of ontological uncertainty \citep{spiegelhalter2017risk}, measuring the limitations of the proposed prediction pipeline as a description of reality. Ontological uncertainty, as opposed to the perhaps more familiar terms of epistemic and aleatoric uncertainty \citep{malinin2018predictive, abdar2021review, hullermeier2021aleatoric}, is a form of predictive uncertainty often overlooked in the machine learning community. In the context of pulmonary nodule detection, we explore conformal prediction to capture ontological uncertainty by making use of hold-out calibration datasets containing nodules with varying levels of consensus. Insights into this topic are meant to aid healthcare regulators in the recent and ongoing process of designing AI governance structures \citep{european2021proposal, food2023drug} by illustrating the risks of using off-the-shelf prediction models in the presence of ontological uncertainty.

\subsection{Summary \& Outline}
\begin{itemize}
    \item We enhance a pulmonary nodule detection model \citep{cardoso2022monai} with conformal risk control (CRC) \citep{angelopoulos2024crc}, leveraging the open-source LIDC-IDRI dataset \citep{armato2011lung, armato2015data}. CRC produces prediction sets with the formal guarantee that the sensitivity of a new CT scan is expected at an arbitrary user-specified level (e.g., 90\%). Among nodules annotated by at least three radiologists, our model achieves 91.35\% sensitivity at a cost of 2.25 false positives per scan. For reference, this sensitivity is higher than generally achieved by individual radiologists, with a slight increase in false positives \citep{armato2009assessment}. 
    \item From a broader perspective, we demonstrate that prediction sets with conformal guarantees are attractive measures of predictive uncertainty in a safety-critical healthcare context, allowing end-users to achieve arbitrary validity by sacrificing more false positives per scan and simultaneously obtaining formal statistical guarantees on model performance. This aids in the challenge of deploying black-box AI models in a large-scale clinical setting. 
    \item Finally, we analyze how consensus among radiologists (i.e., the `ground truth') affects predictive performance, relating ontological uncertainty to the number of false positives in prediction sets. We find lower consensus among radiologists naturally translates into larger prediction sets, to the extent that nodules annotated by only one radiologist cannot be detected efficiently by the underlying prediction model. 
\end{itemize}

\sectionref{sec:case_related} examines related works that studied aspects of the same problem. \sectionref{sec:case_study_meth} covers methodology, starting with the problem setting and followed by a description of the calibration strategies, pairing procedure, dataset and underlying prediction model. \sectionref{sec:case_study_exp} describes the experimental setup and evaluation of results. Finally, \sectionref{sec:case_study_disc} discusses broader implications and contains concluding remarks. 

\subsection{Related Work}\label{sec:case_related}
Some works explored conformal prediction for object detection over the past few years. For example, \citet{timans2024adaptive} leverage conformal prediction to obtain uncertainty intervals with guaranteed class-conditional coverage for bounding boxes in multi-object detection. Furthermore, \citet{andeol2023confident, andeol2024conformal} apply conformal prediction to trustworthy detection of railway signals. Finally, \citet{de2022object} explore conformal prediction for object localization in pedestrian detection. These methods have in common that they apply a conformal procedure to \emph{the area of predicted bounding boxes}, such that these correctly cover the true objects. In contrast, we tune \emph{the detected number of objects} directly. That is, the mentioned methods shrink or enlarge the area of a fixed number of predicted boxes, while we aim to shrink or enlarge the number of predicted objects themselves. Notably, the ultimate goal of limiting undetected objects is similar. In the context of pulmonary nodule detection, we assume that radiologists are mostly interested in a small set of candidates to evaluate, and that they will derive the exact area of individual nodules during further manual inspection. Therefore, it is of primary importance to know the amount of nodules in a particular scan and their rough location, and their exact size is only secondary. To summarize, our objective emphasizes small nodules that might otherwise be missed by object detectors aiming for box coverage.    

A related medical application of the conformal prediction framework include false negative rate control for tumor segmentation \citep{angelopoulos2024crc, bates2021distribution}, where the focus lies on the number of undetected \emph{pixels} instead of the number of undetected \emph{objects}. Furthermore, \citet{angelopoulos2024conformal} explored conformal triage for medical imaging, tuning a binary image classifier to simultaneously control the false negative rate and false positive rate through the Learn Then Test (LTT) \citep{angelopoulos2021learn} method. Our setup differs in that we consider binary classification on the level of anchor boxes, while controlling the false negative rate on the image level through the related but different CRC method. 

CRC differs from other conformal methods in several key aspects. First, it is a strict generalization of split conformal prediction \citep{papadopoulos2002inductive, lei2018distribution} to any monotonic risk function, since split conformal prediction considers only risk functions pertaining to the binary loss function, with the aim of controlling the probability of coverage. Second, the risk-controlling prediction sets (RCPS) method \citep{bates2021distribution} similarly considers monotonic risk functions, but targets more complex guarantees in high probability, instead of guarantees in expectation. Finally, the learn then test (LTT) method \citep{angelopoulos2021learn} targets guarantees in high probability identical to RCPS, but is more generally applicable to any non-monotonic risk function. Relevant to our setup is that in relation to RCPS and LTT, CRC is substantially more sample-efficient, thus leading to considerably smaller prediction sets when only a limited number of calibration samples is available \citep{angelopoulos2024crc}. This is a valuable property in the context of AI-assisted pulmonary nodule detection, since well-annotated CT scans are highly scarce resources.

\section{Methods}\label{sec:case_study_meth}
We treat nodule detection as a multi-dimension binary classification problem. Given a training dataset of 3D CT scans and corresponding set of ground-truth nodule locations, we fit a one-stage object detector $\hat{f}: \set{X}\rightarrow \set{P}(\mathbb{R}^6\times[0,1])$, where $\set{P}$ denotes the power set. The object detector takes as input a CT scan $X\in\set{X}$ and divides it into a large set of overlapping anchor boxes producing a dense grid covering a scan. For example, the object detector creates one or more anchor boxes with varying sizes for each spatial location, and treats each anchor box as a nodule detection task. Then we obtain a set of anchor boxes $\hat{f}(X):=\{(\hat{c}_1, \dots, \hat{c}_6, \hat{p})_j\}^{D(X)}_{j=1}$, with size $D(X)$ depending on the spatial dimensions of the input scan. Each anchor box consists of six real-world location coordinates defining the box $\hat{f}(X)^c_j:=(\hat{c}_1,\dots,\hat{c}_6)_j$, together with a confidence estimate $\hat{f}(X)^p_j:=\hat{p}_j$ indicating whether the box contains a nodule.

Now we assume access to a fresh calibration dataset $\{(X_i,Y_i)\}^n_{i=1}$ to calibrate our prediction model, where each $Y_i\in \set{P}(\mathbb{R}^6)$ represents the set of of true annotated nodules corresponding to a scan $X_i\in\set{X}$. This takes the form of a set $Y_i=\{(c_1, \dots, c_6)_j\}^{O(X_i)}_{j=1}$ with size $O(X_i)$, where $0<O(X_i) \ll D(X_i)$ for all $1\leq i\leq n$. That is, the number of nodules in a scan is positive but tiny compared to the large number of anchor boxes. 

We transform the predicted anchor boxes into a prediction set $C_\lambda(X_{i})$\footnote{Technically, since we estimate $\lambda$ on the calibration dataset, $C_\lambda:\set{X}\rightarrow \set{P}(\mathbb{R}^6)$ is a set-valued random function that maps a scan $X_i\in\set{X}$ to a prediction $Y_i\in\set{P}(\mathbb{R}^6)$, conditional on the pre-trained prediction model $\hat{f}$. However, for simplicity we refer to $C_\lambda(X_i)$ as a prediction set.} through 
\begin{equation}\label{eq:pred_set}
    C_\lambda(X_{i}):=\{\hat{f}(X_{i})^c_j: \hat{f}(X_{i})^p_j\geq \lambda\}.
\end{equation}
The index parameter $0\leq\lambda\leq 1$ indicates the size $\hat{O}(X_i)$ of the prediction set $C_\lambda(X_i)$, where smaller $\lambda$ lead to larger prediction sets. It can be interpreted as a lower bound on the confidence estimate corresponding to an anchor box, filtering out anchor boxes with confidence below $\lambda$. For example, this could be a lower bound on the softmax output of a neural network. 

To find the set of correct anchor boxes in $C_\lambda(X_{i})$, a pairing procedure is required. Let $\text{Pair}(Y_i,C_\lambda(X_{i}))$ denote such a procedure, returning the set of correctly predicted anchor boxes. We defer the specific pairing procedure we employ to \sectionref{sec:case_meth_pair}. Then the number of true positives, false positives and false negatives for a particular scan is denoted as, respectively,
\begin{align*}
    \text{TP}_\lambda(X_i)&:=\text{Pair}(Y_i,C_\lambda(X_{i})),\\
    \text{FP}_\lambda(X_i)&:=|C_\lambda(X_{i})|-\text{TP}_\lambda(X_i), \\
    \text{FN}_\lambda(X_i)&:=|Y_i|-\text{TP}_\lambda(X_i).
\end{align*}
The Fale Negative Rate (FNR) per scan is the expected fraction of false negatives, that is,
\begin{equation}\label{eq:risk_function}
    R^\text{FNR}_{i}(\lambda):=\mathbb{E}[L^\text{FNR}_i(\lambda)]=1-\mathbb{E}\Big[\frac{\text{TP}_\lambda(X_i)}{\text{TP}_\lambda(X_i)+\text{FN}_\lambda(X_i)}\Big],
\end{equation}
where the loss $L^\text{FNR}_i(\lambda)$ is implicitly defined. Note that the FNR is inversely related to the sensitivity per scan, such that minimizing the FNR is equivalent to maximizing the sensitivity per scan. 

We are interested in producing a prediction set for a new scan $X_{n+1}$ that has tight control over the FNR, and thus over the sensitivity per scan. That is, we desire to limit the amount of undetected nodules, while simultaneously obtaining small prediction sets for radiologists to evaluate. To satisfy these requirements, our objective is to tune the parameter $\lambda$.

\subsection{FROC Analysis}\label{sec:case_meth_froc}
Binary classifiers are generally evaluated by radiologists on a held-out dataset in terms of sensitivity and specificity, using familiar receiver operating characteristics (ROC) plots. Then, multiple classifiers are compared using area under the curve (AUC) measures. Each point on such a plot is determined by a value of the confidence threshold $\lambda$, indicating the degree of confidence for a system to distinguish nodules from non-nodules. Setting the threshold amounts to finding a suitable balance between sensitivity and specificity. However, when evaluating models on imbalanced datasets, where the number of negatives significantly outweighs the number of positives, a precision-recall curve (PRC) plot is generally more informative \citep{saito2015precision}. In such cases, the number of negatives significantly outweighs the number of positives, such that the (visual) interpretability of specificity in a ROC plot can be deceptive. This is the case for pulmonary nodule detection, since the number of anchor boxes greatly exceeds the number of nodules per scan. 

If, additionally, the number of positives is not known a priori, such as the total number of nodules present in a CT scan, a free response operating characteristic (FROC) plot can be an appropriate alternative \citep{chakraborty2013brief, miller1969froc, bunch1977free}. It is common practice in AI-assisted pulmonary nodule detection to evaluate model performance using FROC analysis \citep{van2010comparing, gu2021survey}. While a PRC plot shows the \emph{average} sensitivity (equivalent to recall, that is, balancing true positives and false negatives) on the $x$-axis, a FROC plot displays the sensitivity \emph{over all nodules combined} on the $y$-axis. Furthermore, a PRC plot includes precision, while a FROC plot contains the number of false positives. To be precise, 
\begin{align*}\label{eq:metrics}
    \text{False Positives}_\text{FROC}(\lambda)&=\frac{1}{n}\sum^n_{i=1}\text{FP}_\lambda(X_i),\\
    \text{Sensitivity}_\text{FROC}(\lambda)&=\frac{\sum^n_{i=1}\text{TP}_\lambda(X_i)}{\sum^n_{i=1}\text{TP}_\lambda(X_i) + \text{FN}_\lambda(X_i)},\\
    \text{Sensitivity}_\text{PRC}(\lambda)&=\frac{1}{n}\sum^n_{i=1}\frac{\text{TP}_\lambda(X_i)}{\text{TP}_\lambda(X_i) + \text{FN}_\lambda(X_i)}, \\
    \text{Precision}_\text{PRC}(\lambda)&=\frac{1}{n}\sum^n_{i=1}\frac{\text{TP}_\lambda(X_i)}{\text{TP}_\lambda(X_i)+\text{FP}_\lambda(X_i)}. 
\end{align*}
To emphasize the subtle but important difference in the sensitivity metric, consider a dataset with two scans, where an object detector detects 9 out of 10 nodules in the first scan, and 1 out of 2 nodules in the second scan. A FROC plot would display a sensitivity of $10/12\approx 83\%$, while a PRC plot yields $7/10=70\%$. That is, $\text{Sensitivity}_\text{FROC}$ assigns less weight to scans with less nodules. Therefore, if complex cases are concentrated in a small subset of scans, an AI system tuned by FROC analysis fails silently, achieving low sensitivity on scans with complex nodules, while good performance overall. Although this is beneficial to obtain a \emph{useful} system from a radiologist perspective, it is detrimental to obtain a \emph{safe} system from a patient perspective. To circumvent this issue, we interpolate FROC and PRC plots, plotting $\text{Sensitivity}_\text{PRC}$ against $\text{False Positives}_\text{FROC}$ and evaluating performance on these metrics. This is further reflected in our choice of loss function in \equationref{eq:risk_function}, aligned with $\text{Sensitivity}_\text{PRC}$ and deviating from common practice in FROC analysis. 

\subsection{Conformal Risk Control}\label{sec:case_meth_crc}
Conformal risk control (CRC) \citep{angelopoulos2024crc} is a simple method to tune $\lambda$ based on a held-out calibration dataset. To start, we compute the empirical risk $\widehat{R}^\text{FNR}_{n}(\lambda):=\frac{1}{n}\sum^{n}_{i=1}L^\text{FNR}_i(\lambda)$ on the calibration dataset. We note here that the FNR is a monotone function of $\lambda$, since smaller $\lambda$ translate to smaller losses $L_i^\text{FNR}$. This is a key requirement for CRC. Then for a user-specified level $\alpha\in [0,1]$, we choose $\widehat{\lambda}$ such that
\begin{equation}\label{eq:crc}
    \widehat{\lambda} = \inf\{\lambda:\frac{n}{n+1}\widehat{R}_{n}+\frac{1}{n+1}\leq\alpha\}.
\end{equation}
Note that $\widehat{\lambda}$ is close to simply choosing the largest $\lambda$ that results in an empirical risk below $\alpha$, but with a finite-sample correction. Since for moderately sized samples the finite-sample correction is negligible, the practical distinctions between $\widehat{\lambda}$ chosen through CRC and FROC analysis are often minor. Anticipating later results, this is reflected in the comparison in \appendixref{app:comparison}. The main advantage of CRC lies in its principled approach and corresponding formal performance guarantee. 

Our goal is to produce a prediction set for a new scan $X_{n+1}$. The key assumption is that our calibration dataset and the new scan are \emph{exchangeable}, that is, their joint distribution is invariant to any finite permutation. Informally, this means that our calibration data is a representative sample for the new scan $X_{n+1}$. If the calibration dataset and new scan are not only exchangeable but also i.i.d., then CRC is optimal (up to a small factor) compared to alternative methods to estimate $\widehat{\lambda}$ \citep{angelopoulos2024crc}. 

Under the assumption of exchangeability, and choosing $\widehat{\lambda}$ according to \equationref{eq:crc}, the resulting prediction set comes with the following marginal risk control (MRC) guarantee \citep[for a proof see][]{angelopoulos2024crc}. MRC is formulated as an expectation over the randomness in the calibration dataset and the new observation $(X_{n+1}, Y_{n+1})$,
\begin{equation}\label{eq:mrc}
    \mathbb{E}[R^\text{FNR}_{n+1}(\widehat{\lambda}) ]\leq \alpha.
\end{equation}
In plain English, at the chosen level $\widehat{\lambda}$, the FNR of a new scan is expected to be below the target level $\alpha$. For example, at $\alpha=0.1$ we expect to miss at most 10\% of all nodules in a scan, where a more stringent target level $\alpha$ translates to smaller $\widehat{\lambda}$ and thus larger prediction sets $C_\lambda(X_{n+1})$. We remind the reader that the FNR and sensitivity metrics are inversely related, such that FNR control at level $\alpha$ implies sensitivity control at level $1-\alpha$. We further note that the MRC guarantee holds in finite samples and does not require any assumptions on the distribution of our scans and labels, therefore avoiding the introduction of possibly erroneous parametric distributions. 

\subsection{Pairing Procedure}\label{sec:case_meth_pair}

Our $\text{Pair}(Y_i,C_\lambda(X_{i}))$ procedure pairs anchor boxes with a true annotated nodule, returning a subset of correct anchor boxes in the prediction set. We use a variation of Hungarian matching \citep{kuhn1955hungarian} to check if an anchor box matches to a true nodule. For all pair-wise combinations of nodules in $Y_i$ and $C_\lambda(X_{i})$, we calculate the intersection-over-union (IoU). Then, each nodule in $Y_i$ is paired with the single anchor box with the highest confidence level among the anchor boxes with strictly positive $\text{IoU}$. Computationally, such a brute-force pairing procedure is feasible for our dataset size and computational resources (one A100 GPU). For other use-cases than pulmonary nodule detection, fine-tuning of the pairing procedure is required to manage computational resources and mitigate the risks discussed below. 

The matching strategy we employ is conservative, since a high-confidence anchor box only needs to have an arbitrarily small non-zero IoU to be paired with a true nodule. The reason is twofold. First, we want to make sure that there are no true nodules without (potentially) paired anchor box. This would lead to the undesirable situation that the largest possible prediction set does not include all nodules in $Y_i$ and thus the FNR cannot reach zero. Second, we assume that radiologists are mostly interested in the rough location of possible nodules, deriving the exact area of individual nodules during further manual inspection. Empirical checks on the dataset introduced below show that only two nodules cannot be paired with an anchor box by our pairing procedure, but upon further manual inspection this is due to documentation errors in nodule coordinates. 

Generally, for small $\lambda$, each nodule in $Y_i$ overlaps with multiple anchor boxes in $C_\lambda(X_{i})$ due to the dense grid of anchor boxes covering $X_i$. For large $\lambda$, some nodules in $Y_i$ will not be paired with an anchor box, resulting in a higher loss. An implicit assumption in this procedure is that no anchor box is paired with more than one true nodule, which could be the case if an anchor box has a high confidence level and overlaps with multiple true nodules. Although in practice there is no restriction preventing this, we observe empirically that the small amount of true nodules per scan are generally far apart relative to the size of the returned anchor boxes, such that no anchor box overlaps with more than one nodule.

\subsection{LIDC-IDRI Dataset}\label{sec:case_meth_data}
We make use of the LIDC-IDRI dataset, which is the largest publicly available reference database of thoracic CT scans \citep{armato2011lung, armato2015data}. The dataset contains 1,018 clinical and low-dose thoracic CTs, collected from 1,010 lung patients through a collaboration among seven academic institutions and eight medical imaging companies. Since the raw scans in the LIDC-IDRI dataset have various voxel sizes, all scans used in training and calibration are resampled into size $0.703125\times 0.703125 \times 1.25$mm.  

The LIDC-IDRI dataset comes with lesion annotations from four experienced thoracic radiologists following a two-stage scan annotation process. In the first blind-read phase, each radiologist independently reviews each CT scan and assigns regions of interest to one of the categories `nodule $>= 3$mm', `nodule $< 3$mm' or `non-nodule $>= 3$mm'. In the second unblinded-read phase, each radiologist receives the anonymized marks of other radiologists and independently reviews their own findings to come to a final solution. The goal of the annotation process is to find all possible nodules in the available scans without requiring forced consensus. For the purposes of this study, we are interested in the category `nodule $>= 3$mm' and discard the others. Substantial variability exists across radiologists in the task of lung nodule detection in CT scans, even among experienced thoracic radiologists \citep{armato2009assessment}. To illustrate this point, in the LIDC-IDRI dataset, there are 2,669 lesions marked as `nodule $>= 3$mm' by at least one radiologist, of which only 928 were marked by all four radiologists.

The well-known LUNA-16 dataset \citep{setio2017validation} is a subset of the LIDC-IDRI dataset. Of the total 1,018 CTs, scans with a slice thickness greater than 2.5mm are discarded, as well as scans with inconsistent slice spacing or missing slices. This leaves 601 CTs that contain at least one `nodule $>=3$mm' annotated by at least three radiologists. The LUNA-16 dataset contains a total of 1,186 nodules, averaging close to two per scan, but ranging between one and twelve. This dataset is used to construct the pre-trained detection model that we use as black-box predictor in our conformal procedure. 

To realize a non-overlapping calibration and test dataset suitable for our conformal procedure, we slightly relax the constraints used to obtain the LUNA-16 dataset. Starting with the 417 CTs not in the LUNA-16 dataset, we discard scans with a slice thickness greater than 3.0mm (instead of 2.5mm). Subsequently, we construct four calibration and test datasets by filtering out the scans that contain at least one `nodule $>=3$mm' annotated by at least $r$ radiologists, denoted as `Set $r$', where $r\in\{1, 2, 3, 4\}$. Within each of Set $r$, a random split is used to obtain a calibration and test set of equal size. The datasets are summarized in \tableref{tab:lidc_idri}.

\begin{table}[tbp]
\setlength{\tabcolsep}{8pt}
\floatconts
    {tab:lidc_idri}
    {\caption{CT scans in the LIDC-IDRI dataset.}}
    {\begin{tabular}{l|l|llll}
    \toprule
    \multicolumn{6}{c}{\textbf{LIDC-IDRI}}\\ 
    \midrule
    & \textbf{Training} & \multicolumn{4}{c}{\textbf{Calibration/Testing}}\\
    \textbf{Name} & \textbf{LUNA-16} & \textbf{Set 1} & \textbf{Set 2}   & \textbf{Set 3}   & \textbf{Set 4}\\ 
    \midrule
    Consensus (nr. radiologists) & $>=3$ & $>=1$ & $>=2$ & $>=3$ & $>=4$ \\
    Nodule diameter (mm)        & $>= 3.0$     & $>= 3.0$   & $>= 3.0$ & $>= 3.0$ & $>= 3.0$ \\
    Slice thickness (mm) & $<= 2.5$     & $<= 3.0$     & $<= 3.0$    & $<=3.0$    & $<=3.0$    \\ 
    \hline
    Nr. CT scans & 601 & 277 & 197 & 98  & 83 \\
    Nr. nodules     & 1186          & 629           & 381             & 197             & 130            \\ 
    Range of nodules per scan & $(1 - 12)$ & $(1 - 16)$ & $(1 - 13)$ & $(1 - 9)$ & $(1 - 5)$ \\ 
    \bottomrule
    \end{tabular}}
\end{table}

\subsection{Prediction Model}\label{sec:case_meth_model}

We make use of a pre-trained prediction model for volumetric detection of pulmonary nodules from CT scans\footnote{Lung Nodule CT Detection Model (version 0.5.9) by Project-MONAI \citep{cardoso2022monai}.}, trained on the LUNA-16 dataset. Recall that the model $\hat{f}$ inputs a raw CT scan, and returns a dense grid of anchor boxes and corresponding confidence estimates depending on the spatial dimensions of the input scan. In our case, the prediction model is a one-stage CNN based on the RetinaNet architecture \citep{lin2017focal}, producing three anchor boxes per spatial location for each of three pyramid levels. The object detector is able to operate in a single step due to use of the focal loss to mitigate class imbalance. That is, a large number of locations in a scan is evaluated, yet only a small minority contains objects.

In practice, the number of anchor boxes actually returned is conservatively filtered to a smaller number to optimize computational efficiency and reduce the number of false positives in the final prediction sets. In the feature pyramid network of the RetinaNet architecture, per level only the 100k anchor boxes with the highest confidence level are retained. Subsequently, the overall amount of anchor boxes is limited to the 100k anchor boxes with the highest confidence level. Afterwards, non-maximum suppression filtering \citep{bodla2017soft} with threshold 0.22 is used to suppress low-confidence anchor boxes that have high overlap with surrounding boxes. Finally, we use a pre-trained prediction model for volumetric segmentation of body segments\footnote{Wholebody CT Segmentation Model (version 0.1.9) by Project-MONAI \citep{cardoso2022monai}.} to discard any anchor boxes outside the lung. The segmentation model is trained on the TotalSegmentator dataset \citep{wasserthal2023totalsegmentator}. Since no uncertainty is estimated in this last probabilistic step, we take care to be highly conservative. The goal of this filtering procedure is to reduce the number of anchor boxes, while still obtaining a dense grid covering a scan such that every true nodule can be matched to an anchor box through our pairing procedure. This is important since a prediction set $C_\lambda(\cdot)$ with $\lambda=0$ should result in a FNR of zero. Empirical checks show that the set of returned anchor boxes is conservatively large and all true nodules can be paired.

\section{Experiments}\label{sec:case_study_exp}
The setup of our experiments is as follows. For the four datasets `Set $r$' ($r\in\{1, 2, 3, 4\}$) (see \sectionref{sec:case_meth_data}), we randomly split the available scans into a calibration and test dataset of equal size. Subsequently, we deploy the prediction model $\hat{f}$ (see \sectionref{sec:case_meth_model}) to infer anchor boxes on the calibration dataset, such that for each scan we obtain the anchor boxes $\hat{f}(X_i)=\{(\hat{c}_1, \dots, \hat{c}_6, \hat{p})_j\}^{D(X_i)}_{j=1}$, with size $D(X_i)$ depending on the spatial dimensions of the input scan. Using the anchor boxes and our pairing procedure (see \sectionref{sec:case_meth_pair}), we leverage one of the three strategies below to estimate a confidence threshold $\widehat{\lambda}$. Then, we again deploy $\hat{f}$ to infer anchor boxes for each scan in the test set and subsequently obtain prediction sets using $C_{\widehat{\lambda}}(\cdot)$ (\equationref{eq:pred_set}). 

We leverage three straightforward strategies to estimate a confidence threshold $\widehat{\lambda}$.
\begin{itemize}
    \item \textbf{Naive}: We simply use $\widehat{\lambda}=0.5$, independent of the information in the calibration dataset. The value 0.5 is arbitrary, but meant to reflect equal confidence assigned to the nodule and non-nodule class. For well-calibrated prediction models, the Naive strategy might result in reasonable prediction sets, although modern neural networks are known to be poorly calibrated \citep{guo2017calibration}. Nonetheless, we include this strategy to mimic the real-world situation where an off-the-shelf prediction model is deployed without further tuning. 
    \item \textbf{FROC analysis}: We follow the methodology described in \sectionref{sec:case_meth_froc}, estimating $\widehat{\lambda}$ empirically such that we detect $90\%$ of all nodules in the calibration set. 
    \item \textbf{Conformal risk control (CRC)}: We follow the methodology described in \sectionref{sec:case_meth_crc}, estimating $\widehat{\lambda}$ through \equationref{eq:crc} with $\alpha=0.1$, thus targeting $90\%$ sensitivity per scan. The target level is picked for illustration purposes, and in practice, a reasonable trade-off between target sensitivity and FPs per scan should be dependent on the clinical context. 
\end{itemize}

We evaluate the prediction sets produced by each strategy by computing the sensitivity, precision, efficiency (that is, prediction set size), number of false positives and number of false negatives on each scan in the test set, averaging over all scans to obtain an overall metric. To obtain a fair evaluation of Naive and CRC, we run both for $R=10,000$ random splits of the available data, for each of the four datasets. For clarity of the figures shown in the next section, we omit FROC analysis due to its similarity to CRC, and we refer to \appendixref{app:comparison} for figures comparing all three strategies using $R=1,000$ random splits.

\subsection{Evaluation}

\begin{figure}[tbp]
    \floatconts
    {fig:froc_adapted_Validation}
    {\vspace{-5mm}\caption{A plot of the average sensitivity per scan against the average number of false positives per scan, measured on the test data of a random split of Set $r$ ($r\in\{1,2,3,4\}$). The colored dots highlight the confidence thresholds estimated.}}
    {\includegraphics[width=0.8\textwidth]{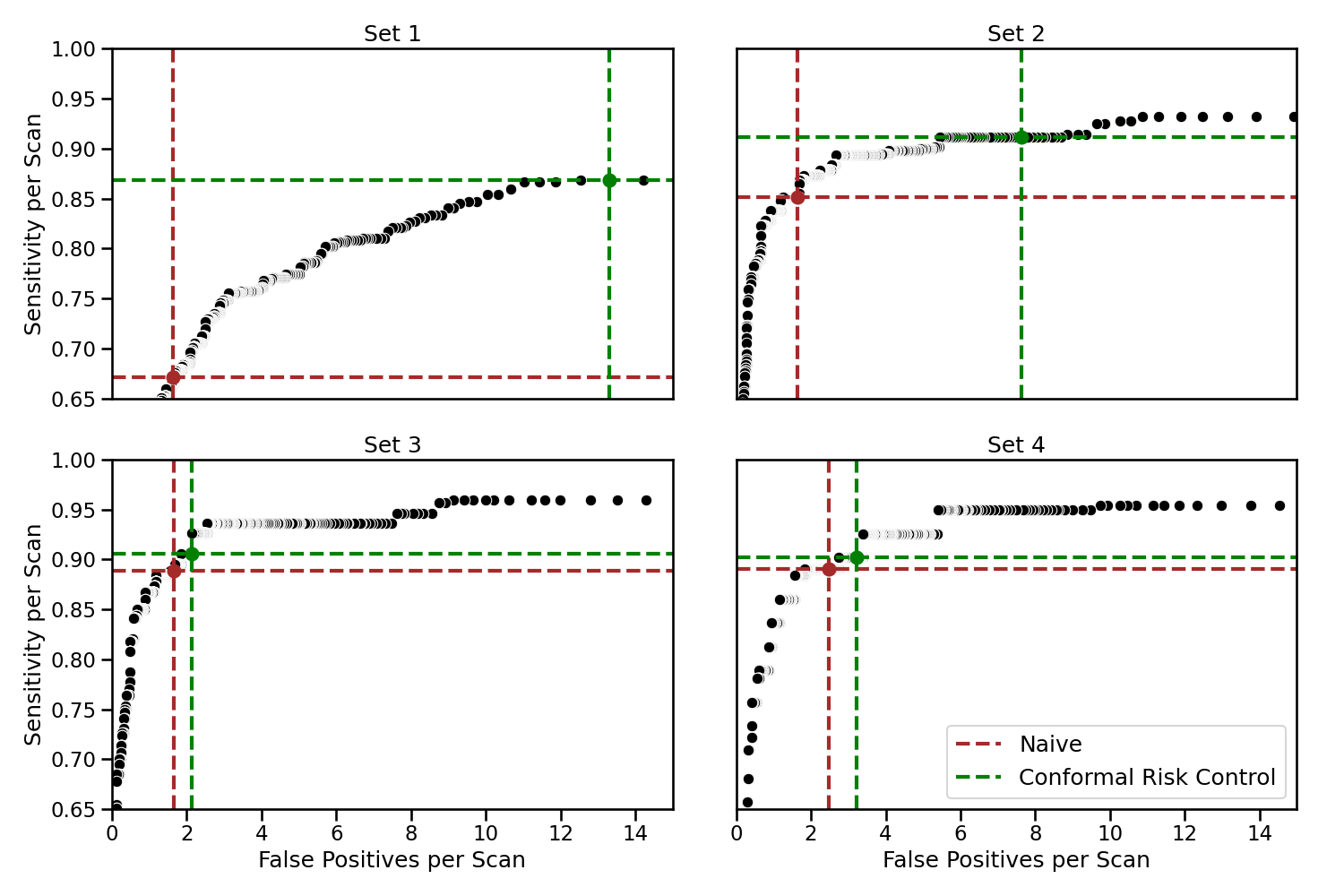}}
\end{figure}

In \figureref{fig:froc_adapted_Validation} we show an adapted FROC plot for a random split of Set $r$ ($r\in\{1,2,3,4\}$) into calibration and test set. The plot is an adaptation of a regular FROC plot since we replace the sensitivity over the set of all nodules with the sensitivity per scan averaged over all scans in the test set (see \sectionref{sec:case_meth_froc} for details). The highlighted points indicate the chosen level of $\widehat{\lambda}$ based on the calibration set. We observe that the Naive strategy undercovers on Set 1 and Set 2, while CRC only slightly undercovers on Set 1. This is in line with the theory outlined in \sectionref{sec:case_meth_crc}, describing that the CRC strategy targets a mean sensitivity of 90\%, yet any particular observed sensitivity is subject to the randomness in the calibration data and a finite test set, and thus may not exactly match the target level. 

The naive strategy is based on $\widehat{\lambda}=0.5$ to mimic the `factory settings' of an off-the-shelve prediction model not subjected to hyperparameter tuning. However, one could argue for other numbers, such as $\widehat{\lambda}=0.9$. \figureref{fig:froc_adapted_Validation} illustrates that larger values of $\widehat{\lambda}$ result in lower sensitivity and less false positives, by construction of the confidence threshold. Hence, choosing $\widehat{\lambda}=0.9$, as compared to $\widehat{\lambda}=0.5$, would increase undercoverage of the Naive strategy and yield detrimental results in terms of sensitivity. Therefore, evaluating performance of CRC against $\widehat{\lambda}=0.5$ is conservative. 

\begin{figure}[tbp]
    \floatconts
    {fig:consensus_analysis}
    {\vspace{-5mm}\caption{Empirical histograms of the performance metrics evaluating the strategies to estimate $\widehat{\lambda}$, measured on the test set over $R=10,000$ random splits of Set $r$ ($r\in\{1,2,3,4\}$) into calibration and test set. Bar heights sum to one.}}
    {\includegraphics[width=\textwidth]{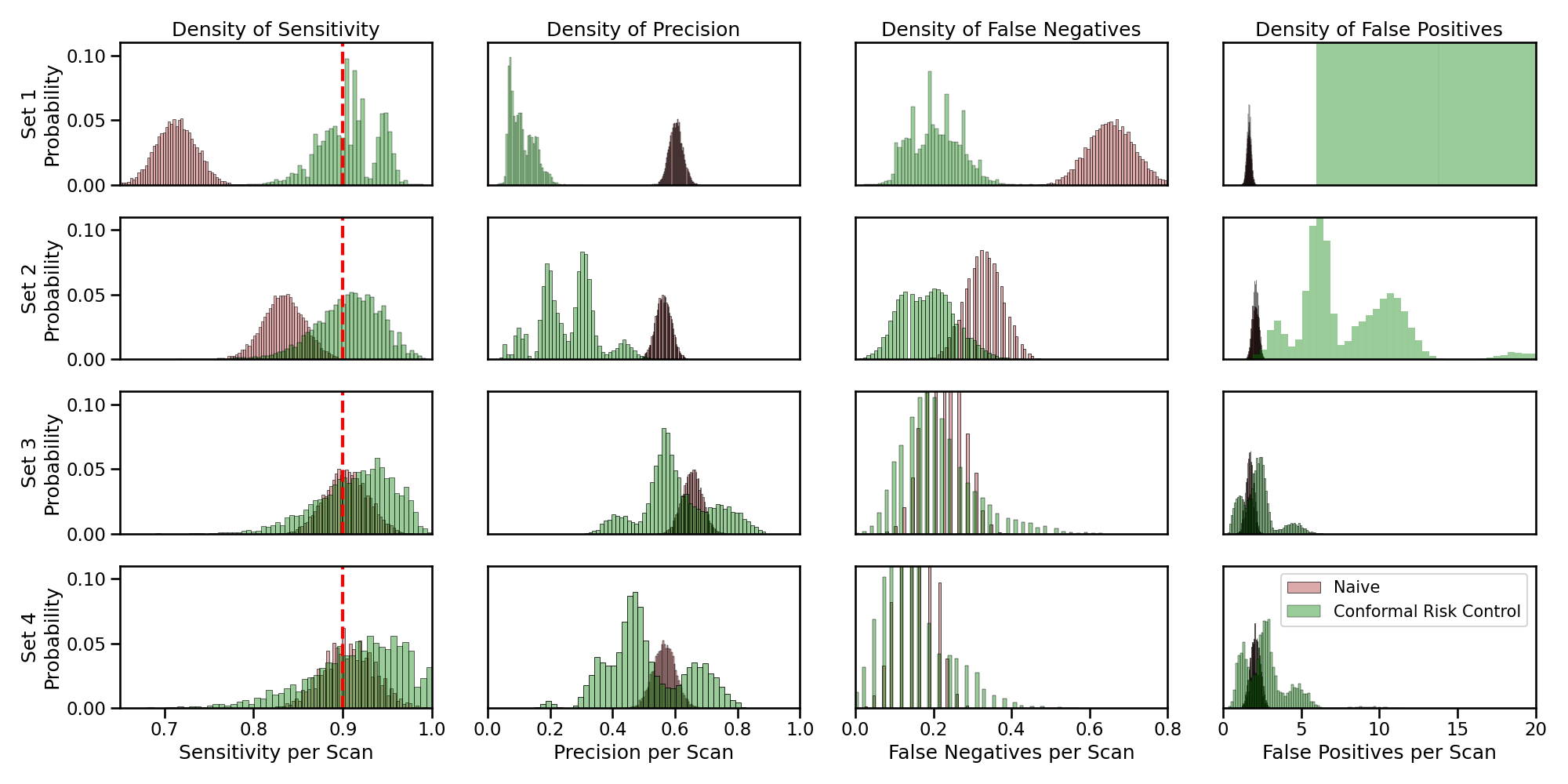}}
\end{figure}

In \figureref{fig:consensus_analysis} we show empirical histograms of relevant performance metrics of $R=10,000$ random data splits. Then \tableref{tab:results} summarizes the empirical mean of the histograms for each of the performance metrics and datasets in \figureref{fig:consensus_analysis}. With regard to the sensitivity, we observe the histogram pertaining to the CRC strategy correctly centers around 0.90 for each of Set $r$, showing CRC is able to obtain the target sensitivity on arbitrary datasets and independent of the connection to the data used in training. Furthermore, the histograms provide intuition behind the MRC guarantee described in \sectionref{sec:case_meth_crc}, connecting the empirical mean of the sensitivity observed in \figureref{fig:consensus_analysis} to the target level $1-\alpha$ in \equationref{eq:mrc}. We further note that the variance of the empirical histogram of sensitivity depends on the randomness in the calibration and test set, where larger sets will lead to empirical distributions more clustered near the mean. 

In terms of ontological uncertainty, we derive from \figureref{fig:consensus_analysis} that a threshold of 0.5 does not suffice to detect nodules annotated by one or two radiologists. This largely depends on the arbitrary threshold, but also on the underlying detection model. The detection model has been trained on the LUNA-16 dataset, which only contains nodules annotated by at least three radiologists. Therefore, the threshold is not calibrated on more ambiguous cases, and applying such an off-the-shelf model clearly requires further tuning. On Set 3 and Set 4, the Naive strategy achieves sensitivity centered at 90\%, likely since the nodules are of similar difficulty to its training data. 

To achieve a sensitivity of 90\% in Set 1 and Set 2, the results of the CRC strategy in \figureref{fig:consensus_analysis} show one should allow for more false positives per scan. This corresponds well to the intuitive approach of representing uncertainty in terms of larger prediction sets. The undesirable consequence of such an approach is that on Set 1, the number of false positives per scan is especially large, well above one hundred, suggesting that deployment in a real-world setting would yield prediction sets without practical benefit for radiologists. Gradual improvements in the capabilities of novel prediction models will decrease the number of false positives per scan, such that substituting a more advanced model in future applications yields smaller prediction sets. 

Finally, we discuss the risk associated with the Naive and CRC strategy. By targeting an arbitrary level of sensitivity, we show in \tableref{tab:results} that the number of false negatives using CRC are brought down to a level of around 1 per 5 scans on average, independent of the dataset used. The level deemed appropriate for clinical deployment of AI systems depends on practical and ethical considerations, but our analysis shows that CRC is a useful methodology to achieve such user-specified levels. Simultaneously, omitting a calibration strategy in clinical deployment introduces possibly silent model failures. We showcase in \tableref{tab:results} that the Naive strategy results in false negatives per scan up to 0.65, meaning more than 1 missed nodule every 2 scans on average, and typically unacceptable in a clinical context.

\begin{table}[tbp]
\setlength{\tabcolsep}{8pt}
\floatconts
    {tab:results}
    {\caption{Performance metrics of the strategies to estimate $\widehat{\lambda}$, measured per scan and averaged over $R=10,000$ random equal splits into calibration and test.}}
    {\vspace{-5mm}\begin{tabular}{ll|lllll}
    \toprule
    \textbf{Dataset} & \multicolumn{1}{c|}{\textbf{Strategy}} & \textbf{Sensitivity} & \textbf{Precision} & \textbf{Efficiency} & \textbf{FN} & \textbf{FP}  \\ 
    \midrule
    & Naive & 0.7129                     & 0.6038 & 3.27   & 0.65 & 1.66    \\
    \multirow{-2}{*}{Set 1} & CRC   & 0.9074                     & 0.1105 & 120.41 & 0.21 & 118.35  \\
                        & Naive & 0.8335                     & 0.5626 & 3.66   & 0.33 & 2.06    \\
    \multirow{-2}{*}{Set 2} & CRC   & 0.9065                     & 0.2596 & 15.97   & 0.19 & 14.24  \\
    & Naive & 0.9034                     & 0.6548 & 3.47   & 0.22 & 1.70    \\
    \multirow{-2}{*}{Set 3} & CRC   & 0.9135                     & 0.6056 & 4.03   & 0.21 & 2.25   \\
                        & Naive & 0.9060                     & 0.5664 & 3.44  & 0.16 & 2.04    \\
    \multirow{-2}{*}{Set 4} & CRC   & 0.9143                     & 0.5098 & 4.15  & 0.15 & 2.75    \\
    \bottomrule
    \end{tabular}}
\end{table}

\section{Discussion}\label{sec:case_study_disc}  
We enhanced an advanced pulmonary nodule detection model with CRC. While individual radiologists generally operate at a slightly lower number of false positives per scan, we demonstrated that an off-the-shelve pulmonary nodule detection model, wrapped in a conformal prediction framework, is competitive in terms of sensitivity, and aided by formal guarantees on model performance. 

From a broader perspective, we illustrated that prediction sets with conformal guarantees are attractive measures of predictive uncertainty in a safety-critical healthcare context, allowing end-users to achieve arbitrary validity by sacrificing efficiency in the form of larger prediction sets. Furthermore, conformal guarantees provide interpretable statements for non-technical radiologists, increasing the potential trust placed in AI-assisted decision-making. Therefore, the conformal prediction framework could aid in the challenge of deploying AI systems in large-scale clinical applications. Specifically when integrated with black-box (that is, neural network) architectures, the reliability guarantees of conformal prediction could serve as a valuable complement to understand the uncertainties surrounding a system's output.

However, patients might seek more fine-grained assurances than those provided by CRC, such as guarantees over individual patient groups, the malignancy of predicted nodules, or the level of distribution shift between observed and future examples. While the former two can be accommodated by known conformal extensions, the latter remains a pitfall in most of the standard conformal prediction literature. Therefore, practitioners should keep in mind that the validity of conformal guarantees breaks down in cases of non-exchangeable data, while solutions require knowledge of the type of drift and are often not obvious. Nowadays, substantial research effort is directed towards, for example, causal machine learning to produce more robust statistical methods for such settings.     

Regarding the risks associated with AI-assisted pulmonary nodule detection, we highlighted that a shift in `ground-truth' consensus between training and test data results in unreliable predictions made by AI systems, indicating that a calibration strategy is required before deploying any off-the-shelf prediction model. This emphasizes the importance of measuring ontological uncertainty in consensus-driven tasks, such as pulmonary nodule detection, as well as the discrepancy between the typical judgement of a healthcare professional and how these judgements are commonly treated in supervised machine learning methods. To increase the mutual understanding between practitioners in AI and healthcare, alignment is needed on how to best encapsulate uncertain truths in the design of AI-based decision support.

\acks{We acknowledge the National Cancer Institute and the Foundation for the National Institutes of Health for their critical role in the creation of the free publicly available LIDC-IDRI dataset \citep{armato2011lung, armato2015data} used in this study. We further acknowledge Project-MONAI \citep{cardoso2022monai} for their efforts in providing a set of open-source frameworks for AI research in medical imaging, such as MONAILabel \citep{DiazPinto2022DeepEdit, DiazPinto2022monailabel}.}

\bibliography{bibliography}

\newpage 

\appendix

\section{Additional Experiments}\label{app:comparison}

\begin{figure}[htbp]
    \floatconts
    {fig:consensus_analysis_froc}
    {\vspace{-5mm}\caption{Empirical histograms of performance metrics of strategies to estimate $\widehat{\lambda}$, evaluated on the test set over $R=1,000$ random splits of Set $r$ ($r\in\{1,2,3,4\}$) into calibration and test set. Bar heights sum to one.}}
    {\includegraphics[width=\textwidth]{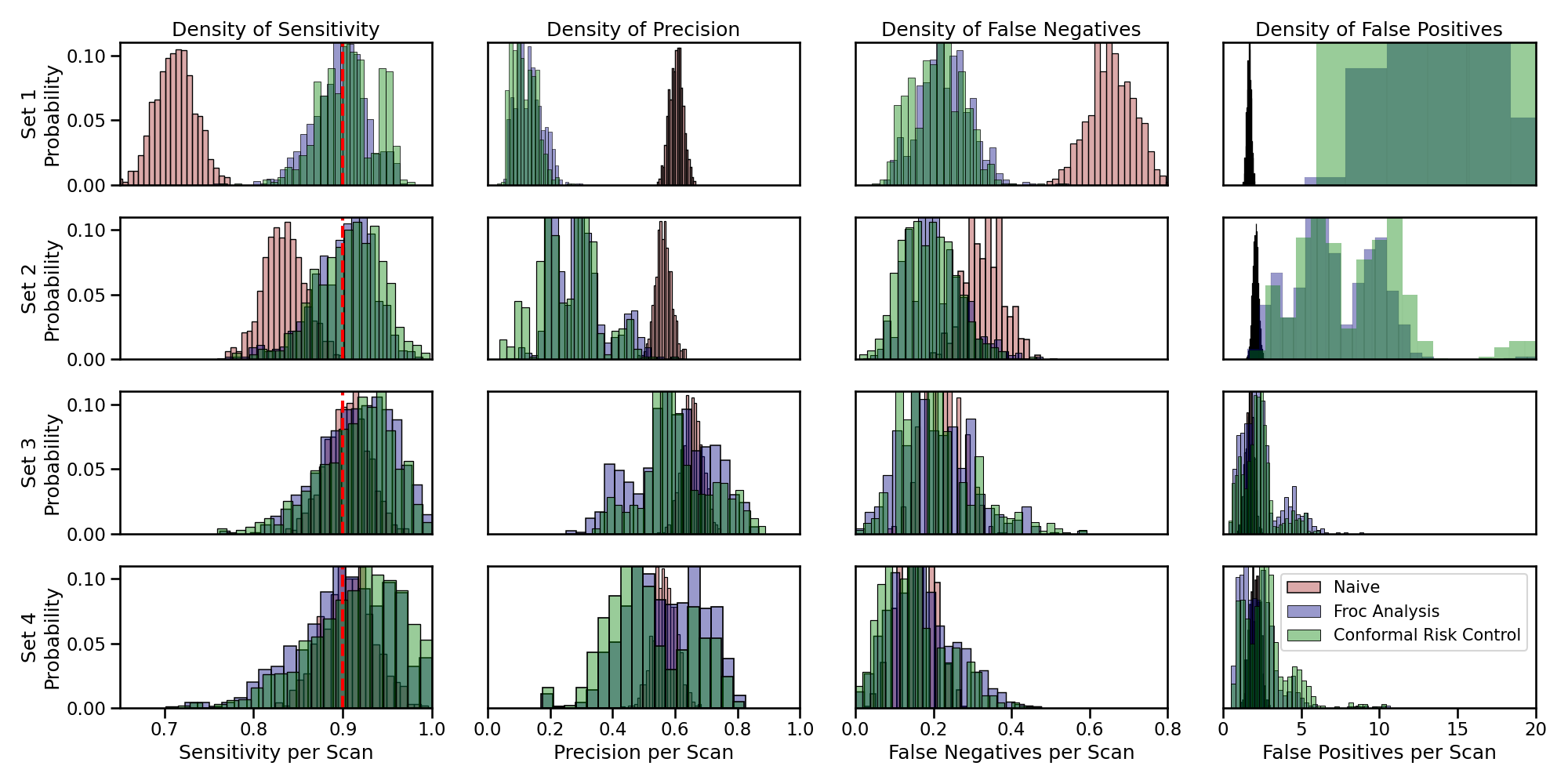}}
\end{figure}

\begin{table}[htbp]
\setlength{\tabcolsep}{8pt}
    \floatconts
    {tab:results_froc}
    {\caption{Performance metrics of strategies to estimate $\widehat{\lambda}$, measured per scan and averaged over $R=1,000$ random equal splits into calibration and test set.}}
    {\vspace{-5mm}\begin{tabular}{ll|lllll}
    \toprule
    \textbf{Dataset} & \multicolumn{1}{c|}{\textbf{Strategy}} & \textbf{Sensitivity} & \textbf{Precision} & \textbf{Efficiency} & \textbf{FN} & \textbf{FP}  \\ 
    \midrule
    & Naive & 0.7129                     & 0.6038 & 3.27  & 0.65 & 1.66    \\
    & FROC  & 0.8971                     & 0.1316 & 48.63  & 0.23 & 46.59   \\
    \multirow{-3}{*}{Set 1} & CRC   & 0.9074                     & 0.1105 & 120.41 & 0.21 & 118.35  \\
                        & Naive & 0.8335                     & 0.5626 & 3.66   & 0.33 & 2.06    \\
                        & FROC  & 0.9005                     & 0.2933 & 9.11   & 0.20 & 7.39    \\
    \multirow{-3}{*}{Set 2} & CRC   & 0.9065                     & 0.2596 & 15.97  & 0.19 & 14.24   \\
    & Naive & 0.9034                     & 0.6548 & 3.47   & 0.22 & 1.70    \\
    & FROC  & 0.9170                     & 0.5945 & 4.18  & 0.20 & 2.39    \\
    \multirow{-3}{*}{Set 3} & CRC   & 0.9135                     & 0.6056 & 4.03   & 0.21 & 2.25   \\
                        & Naive & 0.9060                     & 0.5664 & 3.44   & 0.16 & 2.04    \\
                        & FROC  & 0.9045                     & 0.5652 & 3.63  & 0.17 & 2.24    \\
    \multirow{-3}{*}{Set 4} & CRC   & 0.9143                     & 0.5098 & 4.15   & 0.15 & 2.75    \\
    \bottomrule
    \end{tabular}}
\end{table}

\end{document}